\documentclass{acm_proc_article-sp}

\usepackage[boxed]{algorithm}
\usepackage[noend]{algpseudocode}
\usepackage{placeins}
\usepackage{graphicx}
\usepackage{balance}
\usepackage{color}
\usepackage{subcaption}
\usepackage{epsfig}
\usepackage{hyperref}
\usepackage{epstopdf}

\usepackage{amsmath}
\usepackage{listings}
\usepackage{multirow}
\usepackage{adjustbox}
\usepackage[table]{xcolor}
\usepackage{booktabs,siunitx}
\usepackage{comment}
\usepackage{fixltx2e}
\usepackage{enumitem}
\definecolor{lightgray}{gray}{0.8}

\newcounter{example}[section]
\usepackage{lmodern}

\usepackage[T1]{fontenc}

\usepackage{textcomp}

\numberofauthors{3}

\author{
Jorge Ortiz\\
     \affaddr{IBM Research}
 \alignauthor
  Chien-Chin Huang\\
\affaddr{New York University}
 \alignauthor
 Supriyo Chakaborty \\
 \affaddr{IBM Research}
}

\title{Get More With Less: \\ Near Real-Time Image Clustering on Mobile Phones}

\begin{document}

\maketitle

\begin{abstract}
Machine learning algorithms, in conjunction with user data, hold the promise of
revolutionizing the way we interact with our phones, and indeed their
widespread adoption in the design of apps bear testimony to this promise.
However, currently, the computationally expensive segments of the learning 
pipeline, such as feature extraction and model training, are offloaded to 
the cloud, resulting in an over-reliance on the network and under-utilization of 
computing resources available on mobile platforms. 
In this paper, we show 
%In this paper, we propose
%that by harnessing the computing power \emph{distributed over a number of phones}, 
that by combining the computing power \emph{distributed over a number of phones}, 
judicious optimization choices, and contextual information %we can while implementing the various phases of the 
%learning algorithm, 
it is possible to execute the end-to-end pipeline entirely on the phones at the edge of the network, efficiently.
We also show that by harnessing the power of this combination, it is possible to execute
a computationally expensive pipeline at near real-time.
%phones or based on the current context (such as low battery power, high
%bandwidth) dynamically distribute parts of the pipeline between the phones and the cloud
%and obtain near real-time performance. 
%

To demonstrate our
approach, we implement an end-to-end image-processing pipeline -- that includes feature extraction,
vocabulary learning, vectorization, and image clustering -- on a set of mobile phones.
%it to cluster images distributed across a number of phones. 
Our results show a
75\% improvement over the standard, full pipeline implementation running on the phones without modification --
reducing the time to \emph{one minute} under certain conditions.
We believe that this result is a promising indication that fully distributed, infrastructure-less computing 
is possible on networks of mobile phones; enabling a new class of mobile applications that
are less reliant on the cloud.

%the feasibility in an `infrastructure-less', fully distributed
%computing model is possible

%We are able to  in running time over naive implementations through a
%combination of approximation choices and architectural decisions.
\end{abstract}

\section{Introduction}
Machine learning has revolutionized a broad range of fields and has the potential
to change the way we interact with our mobile phones. The past few years has
seen an increasing number of apps on phones that have exploited learning
algorithms to provide context-aware services such as speaker recognition~\cite{speaker_recog, 
darwin_phones}, activity recognition~\cite{activity_recog}, emotion detection~\cite{emotion_detect},
and several others~\cite{mobile_advert, mobile_human_queue,
mobile_transportation}. However, several machine learning components such as feature
extraction and model training are associated with a high computational and/or communication cost -- consuming a large amount of energy and 
requiring high reliability -- and are typically offloaded to the cloud.
Cloud offloading~\cite{c1,codeoffload_longlast} 
provides several benefits to mobile applications.  It not only 
provides a central location to gather data from \emph{a large number} of phones~\cite{google_maps}, 
enhancing the quality
of the results, but it also extends the battery life of mobile devices~\cite{codeoffload_longlast}.

However, offloading of data also suffers from multiple drawbacks. First, data must be 
transmitted from the device to the cloud, potentially exposing a user's personal information, such as 
location traces or images.  Second, depending on the status of the network connection
the cloud-based service can become unavailable. Such network based outages are
commonplace when many phones are co-located in a fixed geographic region, for example
at stadium or an event in a park. The performance of a cellular connection
depends on the number of users active in your ``cell'' -- diminishing in quality as the service 
cell becomes congested~\cite{cell_perf}. Finally, there is disproportionate
growth in data generation and additional network bandwidth.
It is projected that global mobile data traffic will increase nearly tenfold between 2014 and 2019.
While the 4G/LTE cellular
network can increase its bandwidth by 20x, the projected demand will still exceed capacity in 2016.  In response, many users offload computation through different channels to the internet.
By 2016, more than half of all traffic from mobile devices (almost 14 exabytes) will be offloaded to the fixed network by means of Wi-Fi devices and femtocells each month~\cite{cisco_projection}. 

We propose exploration within a different operating regime -- 
one in which a cellular link is poor or intermittently available and 
users are incentivised to cooperate in order to send useful information to the cloud.
%perform distributed
%computation, and reduce the amount of information being sent back to the cloud
%resulting in better usage of the existing cellular connection.
For example, if users wish to send images from a popular protest, it would be useful
for their phones to cooperatively choose which images to transmit and the order
in which to transmit them, in order to
%in which the selected images are to be transmitted to 
make the best use of the intermittent cellular link to the cloud.
%poor 
%connectivity (and to allow news outlets to obtain important, real-time information).
Consider the approach when a good link is available.
Phones could individually send their images to a server where 
a common \emph{basis} space (for representing the images) is computed.  The basis is used to then 
compress images into a bag-of-words (BoW) representation.  Each
BoW vector can be used to partition the images into $k$ cluster using K-Means (Lloyd's algorithm)~\cite{kmeans}.
Although this is just one of many ways to cluster images, it is a commonly used pipeline for image clustering~\cite{common_pipeline}.
However, in the new operating regime we consider, the network communication cost dominates.  Therefore, we must execute
this pipeline in the network itself, among the mobile phones.
%the execution of this 
%In the context of our operating regime, however, the stages of this pipline would need to executed 
%entirely in the mobile network.
In this paper, we investigate the key question regarding the feasibility of the
above proposal.  

Cluster computing with mobile phones has been discussed
in the past.  The HPC community has envisioned scenarios where spare compute cycles can be 
used in a mobile cluster~\cite{pocket_cluster}.  With power and communication being a concern, it is clear that not all algorithms
would work well in this setting. However, with the increasing capabilities and compute power of phones, we
believe it is time to revisit this question and examine it more thoroughly.
For deeper analysis, we implement a distributed
clustering application on phones, whereby the phones in the network collaborate to construct a representative 
subset of images to send back to a central repository. We show that a direct implementation of the 
cloud-based version yields very poor results but that several optimization makes the solution feasible.
We achieve over a 4.3x improvement over the naive implementation with over \emph{75\% 
cluster 
similarity} with the full solution. We choose distributed image clustering as it
is computationally expensive and data intensive providing benefits when the
available bandwidth is low or unstable. The algorithm runs in stages with
components whose runtime can be improved with contextual information, further
providing opportunities for optimization and distribution.
We also explore the design space by constructing a simple model
based on the empirical measurements made on the implemented system and observe that dynamic pipeline shifting
as a function of available bandwidth could provide improved performance as we move from a 
network-bound operating regime to a compute-bound one. 

We believe that collaborative computation on mobile phones is under-explored and that there
is an opportunity to leverage \emph{high-density co-location} of mobile phones to
support a new class of applications. We also believe that with the increasing power of mobile phones,
now is the time to consider architectures and techniques that were previously unfeasible -- even just a couple of 
years ago.
Through our initial exploration, we show
that by combining approximation algorithms, distributed computing, and contextual information, we 
can harness the power of machine learning at the edge of the network while reducing the computational overhead.
In the rest of the paper we will describe our implementation, optimizations, and modeling results.  We close
with a discussion about their implication.

%of an image processing pipeline that we run
%on a set of mobile phones.  We show that by combining several optimizations, we can reduce 
%the execution time substantially, without affecting the results of the process by much.  We also discuss 
%the implications of our results and present a model that suggests that an adaptive pipeline allocation mechanism
%could provide performance benefits for similar applications.

\begin{figure*}[ht!]
\centering
        \includegraphics[width=\textwidth]{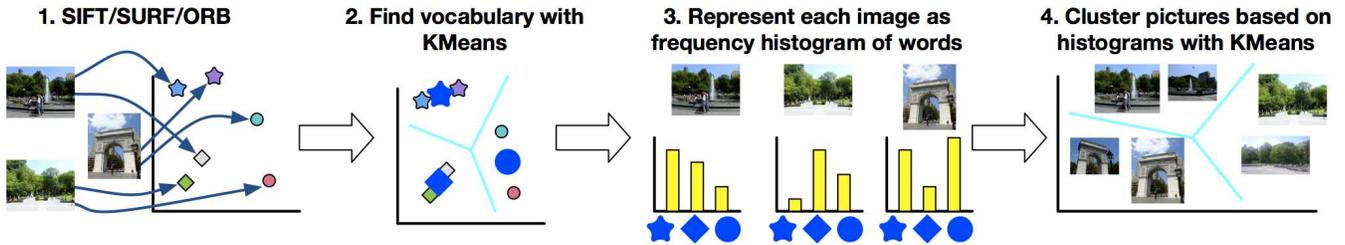}
                \caption{Stages in the machine learning pipeline.  First we run SIFT to extract features of the images, K-Means to construct
                the feature vocabulary for vectorization, we vectorize the images, and we run K-Means on those to cluster the images.  Finally, the
                phones with images closest to each centroid uploads that image.}
                \label{fig:pipeline}
\end{figure*}

\section{Image Clustering Pipeline}

Our image clustering pipeline is illustrated in Figure~\ref{fig:pipeline}.  It consists of four
stages.
%We implement a full image clustering pipeline which is made up of four stages.  
The first stage
runs scale-invariant feature transform (SIFT)~\cite{sift} on each image to extract a set of features.  The features from each image
are treated as a single collection and passed to the next stage.  The second
stage runs K-Means on the collection and treats the \emph{cluster heads} (centroids) as
the representative `words' or vocabulary of the dataset.  The third stage labels the features for each image
and generates a vector, where each row represents a specific word and the corresponding value is the number of occurrences
of that word in the image.
%The third stage looks at the feature set for each
%individual image and classifies each feature using the clustering from the previous step.  In this stage we keep
%count of the number of times each feature is found and vectorize each image into a word-count vector.
This is known as a bag-of-words presentation (BoW)~\cite{fei2005bayesian}.  Finally, each BoW vector is clustered again using
K-Means.  The clusters here represent the different `types' of images in the data set.  In our application, the image closest
to the centroid for each cluster is transmitted to the cloud for storage.  We consider the transmitted image set to be 
representative of the images in the network.
%The pipeline is
%shown in Figure~\ref{fig:pipeline}.

%We use this pipeline throughout our analysis and model the cost of running each component of the pipeline between the
%cloud and the mobile network(Section ~\ref{sec:pshift}). 
%The stages of the pipeline have different cost in the cloud
%and the mobile network.  We measure the overhead in both settings and parameterize the model before sweeping
%through multiple bandwidth and configuration settings.  In this section we will give a brief overview of
%the K-Means algorithm and some initial measurements of the naive implementation in a set of mobile phones.
%We will then go through each optimization and show how it improves the overall performance of the pipeline
%in the mobile network.

\subsection{K-Means Algorithm}
Given a set of $n$ points $\{ x_{0}, x_{1},\dots,x_{n-1} \}$ and $k$ cluster heads (centroids)
$\{ c_{0},c_{1},\dots, c_{k-1} \}$
the K-Means algorithm's goal is to minimize the following function:
\vspace{-0.5mm}
\begin{equation}
\label{eq_tx}
J(X,C) = \sum_{j=1}^{k}\sum_{x_{i}\epsilon C_{j}}{} \parallel x_{i}-c_{j} \parallel^{2}
\end{equation}
\vspace{-0.3 mm}
This function is the square of the distance from each point within a cluster to the cluster's centroid, 
$c_{j}=\frac{1}{\mid C_{j} \mid}\sum_{x_{i}\epsilon C_{j}} x_{i}$.  Minimizing this objective function is
NP-hard~\cite{kmeans_nphard}.  However, the K-Means algorithm is a heuristic algorithm that is guaranteed
to find a local minimum.  The algorithm works as follows: given an initial value for $k$ and a
set of $k$ random coordinates $\{ c_{0}^{0},c_{1}^{0},\dots, c_{k-1}^{0} \}$ the algorithm consists of
the following steps:

\vspace{-2 mm}

\begin{enumerate}
\item \textbf{Assignment step}  Given the current set of centroids, each point 
 $\{ x_{0}, x_{1},\dots,x_{n-1} \}$ is assigned to the corresponding cluster for which $x_{i}-c_{j}$ is smallest.

\item \textbf{Update step} Once each point is assigned to a cluster, recalculate the centroid coordinates
by averaging all the points in a given cluster, $c_{j}=\frac{1}{\mid C_{j} \mid}\sum_{x_{i}\epsilon C_{j}} x_{i}$.
\end{enumerate}

The first step is $O(ndk)$ and the second step is $O(n)$.  The entire algorithm runs in $O(nkdi)$ where
$n$ is the number of points, $k$ is the number of clusters, $d$ is the dimensionality of $x$, and $i$
is the number of iterations that the algorithm takes to converge.  Convergence is defined with a stoppage
criteria which is either 1) a fixed number of iterations, 2) no change in the cluster heads, or 3) a change
in the cluster heads within some $\epsilon$ from one iteration to the next.

% explanation of how to distribute the work in a cluster
In our experiments, we run a distributed version of this algorithm.  In distributed K-Means~\cite{distr_kmeans},
a typical setup contains one master node that manages the iterations of the algorithms and determines when the process is
complete.  The master node broadcasts the starting centroids.  Each node then runs the \emph{assignment step} and
the \emph{update step} locally and reports their new centroid calculation back to the master.  When the master
receives the centroids from each of the nodes, it computes the average of each centroid and initiates another iteration.
The process stops when the stoppage criteria is met.

\subsection{Image Feature Extraction}
For image feature extraction we use the SIFT technique that scans an image, looking for distinctive local features -- present over a fixed
location within the image itself.  These typically indicate fluctuations in pixel values, like
those found through corner-detection techniques.  SIFT is robust to several kinds of transformations, such as scaling,
rotation, affine, 3D perspective and various others.  SIFT outputs a high-dimensional
vector for each feature, known as a descriptor.

Because of the robustness of SIFT, it is typically used to look for similar features across images
that contain overlapping scenes.  Although feature extraction is robust to various kinds of transformations
it still varies enough that similar features across pairs of images from different angles can vary slightly.
To smooth out the variability we cluster them and use the centroids
as the representative feature set.  In our experiments, we tried several other feature-extracting techniques, such as
SURF~\cite{surf} and ORB~\cite{orb} -- with ORB giving us a measurable performance improvement.
Indeed, either algorithm can be substituted for SIFT.  However, in practice, both give worse clustering results.
It is a fundamental tradeoff for execution time improvement.
%We therefore chose SIFT as the image feature extraction algorithm for the pipelines.
%It uses the same floating number
%descriptor structure. Thus it can be easily substituded for SIFT.
%ORB, on the other hand, uses a binary descriptor structure which is more compact and faster for computation.
%Using binary K-Means\cite{bkmeans} as the clustering algorithm, ORB can also be used in the pipelines.
%Among the three algorithms, SIFT generated the best clustering results in our experiments.
%Moreover, image feature extraction is not the main bottleneck in the original pipeline\label{fig:perf_initial}.
%We therefore chose SIFT as the image feature extraction algorithm for the pipelines.

%\subsection{Two Seperate K-Means in Pipeline}
%Images as a Bag of Words
%Cluster of BoW vectors

\subsection{Full Pipeline, In-Network Performance}
We run the entire pipeline, illustrated in Figure~\ref{fig:pipeline}, on a set of three mobile phones: Google Nexus 5, Motorola Nexus 6, and the
Samsung Galaxy S4.
The Nexus 5 has a Quad-core processor 2260 MHz Krait 400 with 32GB of storage.  The Nexus 6 has a Quad-core 2.7 GHz Krait 450
also with 32GB of storage.  The Samsung Galaxy S4 has a Quad-core 1900 MHz Krait 300 and 64 GB of storage.
We collected a set of photos from each of the phones using different resolutions within a park in downtown Manhattan.  We tried to collect a dataset
that approximates one collected from a set of co-located phones.
% specific image size details.  Resolution, size in bytes, pre-processing, etc.

We make a slight modification to a common processing pipeline used for classifying images~\cite{common_pipeline}.
The difference is in the last stage.  Typically, a classifier such as Naive Bayes or SVM is used to classify the photos after they have been vectorized.
We choose to cluster them instead, since our application aims to construct a representative set of photos.  Our approach
clusters the images and transmits the images closest to the centroids.  The setting for $k$ in both k-meeans (in the pipeline) was chosen experimentally.  We
visually assessed the quality of the output before choosing them.  For the purposes of this study, we believe this is acceptable.  Some amount of
pre-processing must be done before running the pipeline in the network.

% opportunities for optimizing k-means
\begin{figure}[h!]
                \centering
		\includegraphics[width=0.45\textwidth]{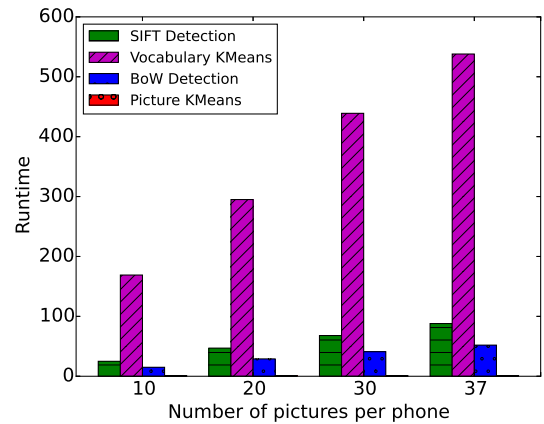}%{./figs/kmeans2_bottleneck}
                \caption{The figure shows that time is takes to execute each stage in the pipeline.  Notice, the first K-Means stage -- to construct the vocabulary from the
                extracted features -- is the main bottleneck in overall performance.}
                \label{fig:perf_initial}
\end{figure}

Figure~\ref{fig:perf_initial}  shows the execution time of each stage in the pipeline for a different number of images.  `SIFT Detection' denotes the SIFT computation stage
which extracts the features from each image, `Vocabulary K-Means' is the K-Means stage that is used to discover a representative set of features/words, 'BoW detection' is the vectorization
step, and 'Picture K-Means' clusters the images using their vector representation.  Note that for all stages, the `Vocabulary K-Means' is by far the most expensive.
With 37 images per phones, the entire pipeline takes an average 679 seconds (over 11 minutes) to complete; the second stage itself takes over eight minutes.

\begin{figure*}[t!]
    \centering
    \begin{subfigure}[b]{0.32\textwidth}
        \includegraphics[width=\textwidth]{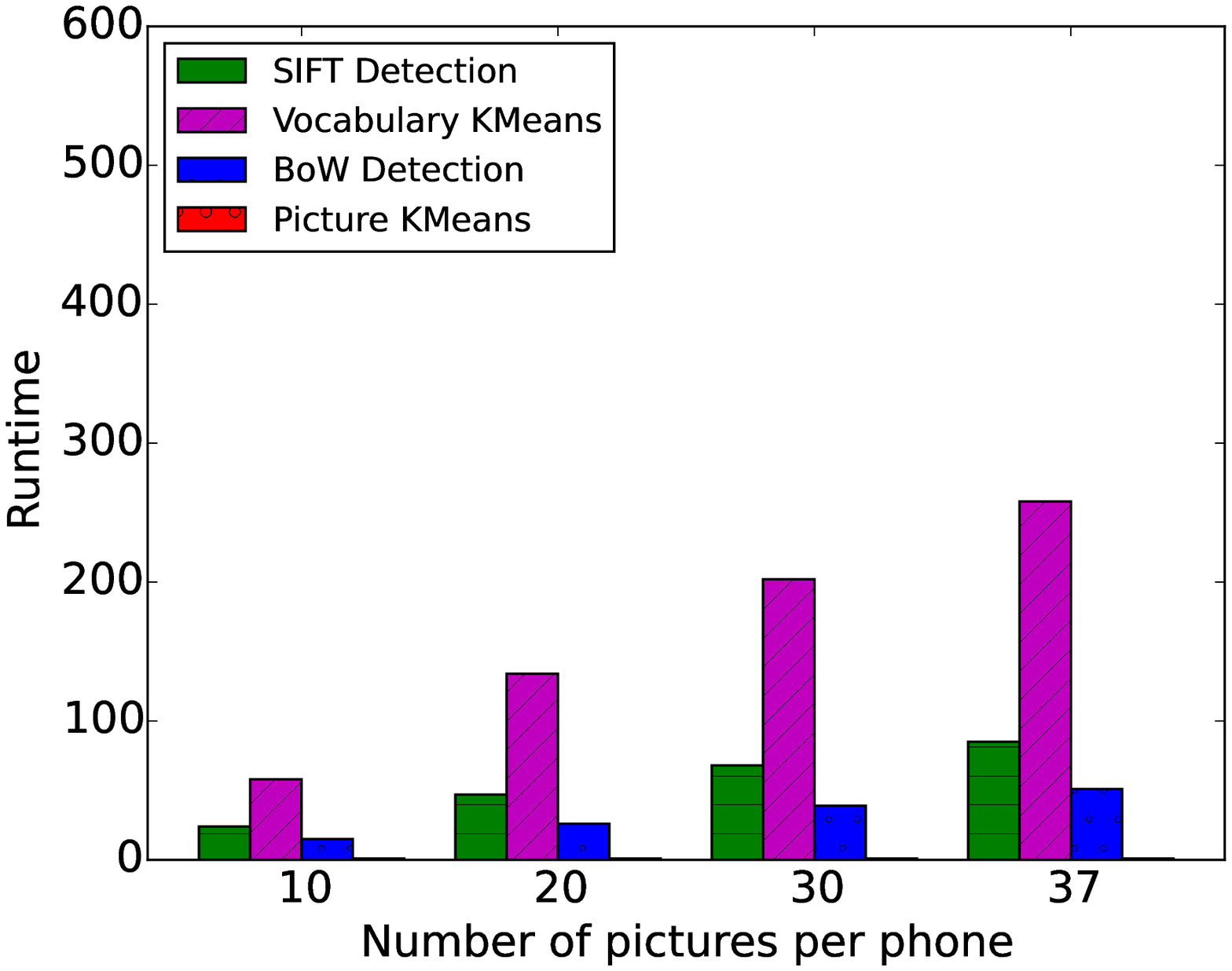}
        \caption{Approx. K-Means with BoW.}
        \label{fig:approx}
    \end{subfigure}
    ~ %add desired spacing between images, e. g. ~, \quad, \qquad, \hfill etc.
    \begin{subfigure}[b]{0.32\textwidth}
        \includegraphics[width=\textwidth]{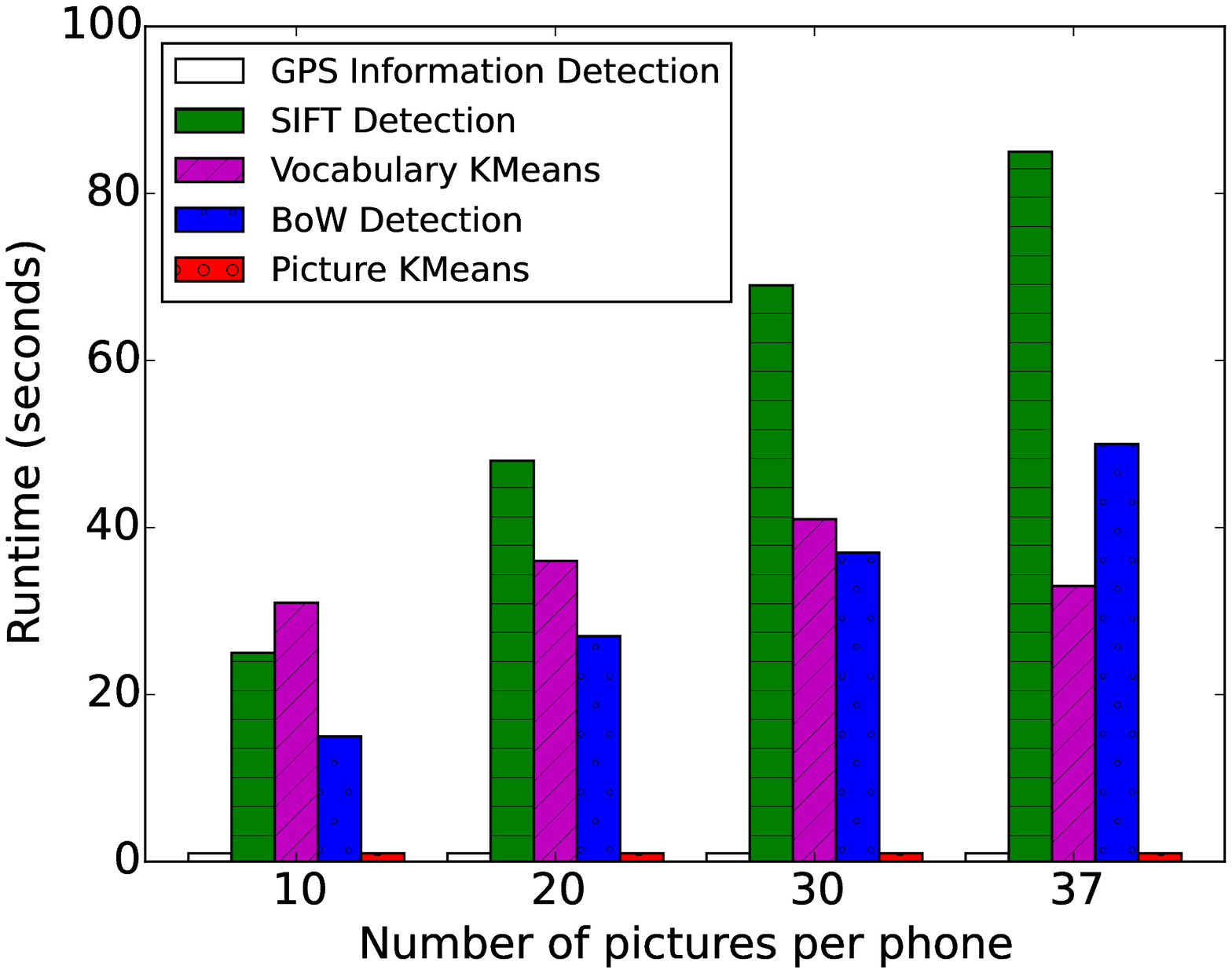}
        \caption{EXIF with BoW.}
        \label{fig:exif_bow}
    \end{subfigure}
    ~ %add desired spacing between images, e. g. ~, \quad, \qquad, \hfill etc.
    \begin{subfigure}[b]{0.32\textwidth}
        \includegraphics[width=\textwidth]{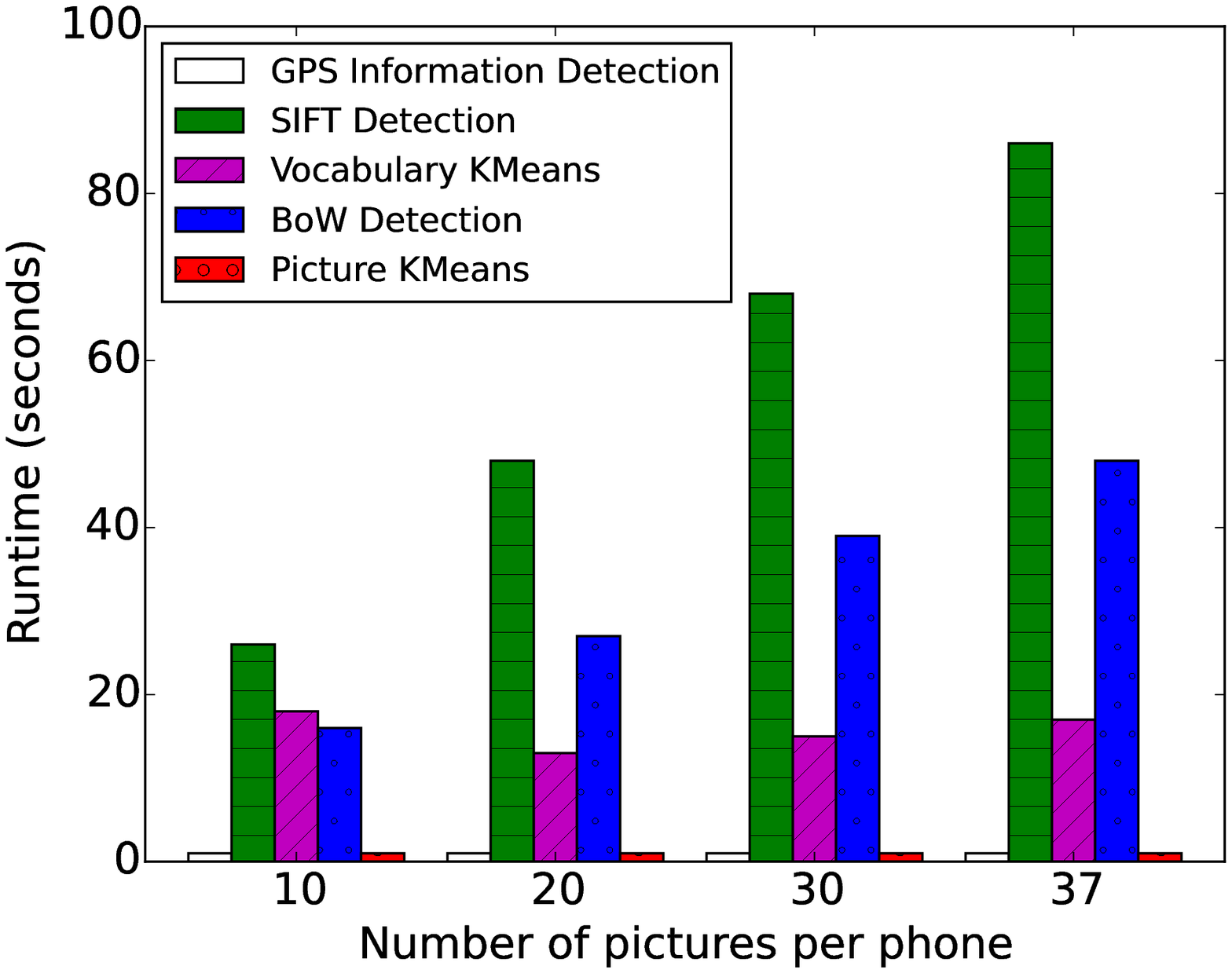}
        \caption{Approx. K-Means with EXIF and BoW.}
        \label{fig:exif_bow_approx}
    \end{subfigure}
    \caption{Performance for each pipeline stage for each combination of optimizations.  The optimizations include context-based seeding with GPS
and orientation data as well as a version of approximate K-Means that discards non-active points in the update step.}\label{fig:optimizations}
\end{figure*}

%\begin{figure}[h!]
%                \centering
%		\includegraphics[width=0.45\textwidth]{./figs/approximate.eps}
%                \caption{Approximate k-means performance break-down of each pipeline stage for a different number of images per phone.}
%                \label{fig:approx}
%\end{figure}
%
%\begin{figure}[h!]
%                \centering
%		\includegraphics[width=0.45\textwidth]{./figs/exif_bow.eps}
%                \caption{EXIF BoW.}
%                \label{fig:exif_bow}
%\end{figure}
%
% \begin{figure}[h!]
%                \centering
%		\includegraphics[width=0.45\textwidth]{./figs/exif_bow_approximate.eps}
%                \caption{EXIF BoW Approximate.}
%                \label{fig:exif_bow_approx}
%\end{figure}             

\subsection{Optimizations}
We implement a number of optimizations in order to decrease the execution time.  The main challenge is maintaining clustering quality
as we add more approximations.  To measure the tradeoff, we compare our optimized pipeline cluster to the original one.
Since the performance bottleneck is in the second stage of processing pipeline, we focus our attention
on improving the execution time of K-Means.  K-Means can be optimized a number of ways.  Recall that the runtime of K-Means
is O(ndki), where n is the number of points, d is their dimensionality, k is the number of clusters, and i is the number of iterations.
We can optimize along any of these four parameters.  %We can minimize the number of points, the dimensionality
%of the points, decrease the number of clusters, and minimize the number of iterations.  
In our work, we focus on
the number of points and the number of iterations.

\textbf {NDK vs SDK}
We implement our pipeline using the Android NDK~\cite{android_NDK}.
The Android NDK allows you to write native code that runs
on the device.  It is typically used by applications that run computationally intensive jobs, providing considerable performance
gains.  We used the NDK to write both our image processing code as well as our clustering code.  Indeed, the performance gains are
significant.  Using 37 images per phone, on three phones, the completion time with the SDK is over two hours.  %We killed the processes
%before it finished.  
The same pipeline, implemented with the NDK finishes in 679 seconds.  Figure~\ref{fig:perf_initial}
shows the execution time of each pipeline stage with the NDK implementation.  Notice that the second stage of the pipeline dominates
the overall execution time.  %In order to drive the execution time down further, we focus our attention on pipeline stages.

%The main optimization necessary is the implementation of the pipeline using the native toolkit for Android~\cite{android_NDK}.
%The native toolkit allows us to implement our algorithm and image processing in C++, which gives us better performance than the
%Android SDK.  The NDK gives us a significant performance improvement over the SDK, as expect.  Moreoever, image processing is
%significantly faster with the NDK.

%Table~\ref{tab:ndk} shows the performance improvement we see between the SDK and the NDK.

% add table

% expand discussion on what was implemented and why it improves the performance

\textbf {Approximate K-Means}
There is a large body of work on approximate K-Means algorithms.  Wang et al.~\cite{kmapprox} observe that in many datasets most of the points
in the \emph{assignment step} do not get re-assigned.  The points that are most likely to get re-assigned are those that sit on the boundary
between any cluster pair.  They refer to these points as \emph{active points}.  Let $d(x_{p},c_{j})$ define the distance between point $x_{p}$
and centroid $c_{j}$.  Also, let $r = 1 - \frac{d(x_{p},c_{i})}{d(x_{p},c_{j})}$ define the relative ratio between the distance between the two closest
centroid $c_{i}$ and $c_{j}$.  They show that active points -- across a set of disparate data sets -- have a distribution whereby over 90\% of active
points have an $r$-value < $0.15$.

\begin{figure}[h!]
                \centering
		\includegraphics[width=0.35\textwidth]{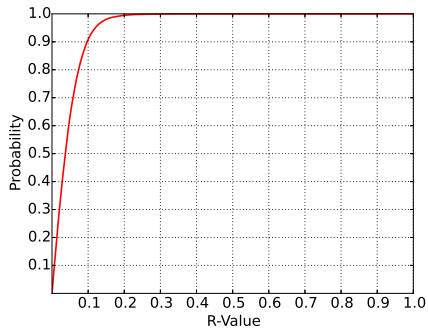}%{./figs/kmeans2_bottleneck}
                \caption{The $r$-value CDF for active points.  Note, non-active points have an $r$-value of 1 and are no plotted in the figure.
90\% of the active points in our dataset have an $r$-value 0.1 or less -- similar to the distribution by Zeng et al.~\cite{kmapprox}.}
                \label{fig:apdist}
\end{figure}

We calculate the $r$-value distribution in our dataset.  Figure~\ref{fig:apdist} shows a CDF of the $r$-value distribution for our \emph{active points}.
 In our implementation, we identify the set of \emph{active points} as the points that change cluster membership from the first iteration to the second one. 
In our dataset, 70\%
of points are considered \emph{active points}; 90\% of which have an $r$-value < $0.10$.  This distribution is similar to the one found in~\cite{kmapprox}.
We discard inactive point in the remaining iterations to reduce the runtime of k-means further.
%We implement our simplified approximation by discarded \emph{inactive} points after the first iteration.  
With 37 photos, the completion
time is 395 seconds, a 41\% reduction from the original pipeline.  Figure \ref{fig:approx}
shows the performance improvement for each stage of the pipeline when approximate K-Means is used.  The average reduction is 45\% for each experiment,
as we vary the number of photos being processed.

\textbf {Metadata Seeding}
Another useful approach to improve K-Means performance is to provide a good set of initial centroids.  Our hypothesis
is that location and orientation can be used to hint about which features may appear in the images.  We recorded GPS and orientation in JPEG
EXIF~\cite{exif} metadata and ran K-Means on this data first.  Then we used the centroid to seed to `Vocabulary K-Means' pipeline stage.
The EXIF K-Means is very inexpensive to run, as $d$ is very small and $i$ is also generally small (it converges quickly).
%We also used the the coordinates of the clusters formed with the metadata to seed each
%of the two K-Means stages in the pipeline. 
We find that seeding the vocabulary-constructing K-Means stage (stage 2) significantly reduces the number of
iterations and improves cluster quality.  It also outperforms the original pipeline when random centroids are used.
Figure~\ref{fig:exif_bow} shows the results of metadata seeding.  We see an overall average reduction of 70\% in execution time from the original
pipeline.  We include the metadata seeding K-Means in our calculation.
%with over 75\% overlap with the clusters formed with a `good' hint.

Our original pipeline performs worse with a random set of centroids, on average, than with a hint from the metadata.  In order to measure the improvement seen with the metadata, we
compare the cluster overlap between a well-seeded full pipeline and one that is seeded with the metadata.  To seed the original pipeline we use a prior run of that k-means step
as our initial seed.
We find that in over 100 runs, there is an
average overlap of 74.7\% with the well-seeded one.  By comparison, the original pipeline with a random centroid seed overlaps with the well-seeded version
by 69.3\% on average.  For our dataset, the 5\% improvement is qualitatively observable.

% table of the different combinations we tried

% percent and absolute improvement figure/table

\textbf{Fully Optimized}

\begin{figure}[h!]
                \centering
		\includegraphics[width=0.45\textwidth]{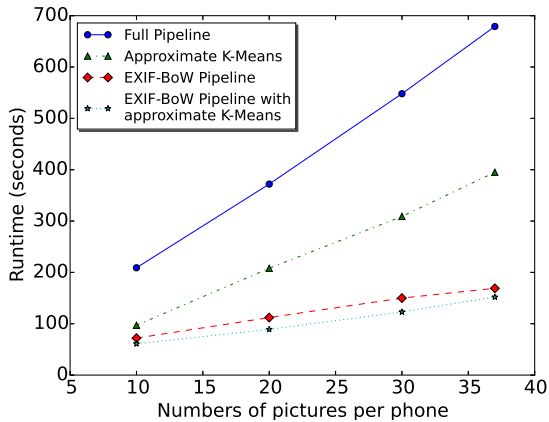}%{./figs/perf_initial}
                \caption{Performance measurements of entire pipelines on three phones.  The execution time of the pipeline increases with the size of the input.}
                \label{fig:perf_total}
\end{figure}

In our fully optimized pipeline, we combine both approximate K-Means and metadata seeding.  Figure~\ref{fig:exif_bow_approx} shows the performance of each stage in
the pipeline.  The overall execution time for 37 photos is 152 seconds.  The average performance reduction from the original is 75\%.  The largest performance improvement
is seen by metadata seeding.  Approximation adds another 5\% reduction.  With only 10 images, the overall execution time is 61 seconds.  The entire pipeline
can run in the mobile network and complete in \emph{one minute}.  We believe this makes it feasible and cheap enough to run as a background process for a family
of mobile sensing applications, without interfering with phone interaction.  Figure~\ref{fig:perf_total} shows a summary of each of the optimization and the original pipeline.
We can see that the biggest performance improvement is seen with metadata seeding.  We can also observe that the optimized pipeline scales better than
the original.  Finally, notice in Figure~\ref{fig:exif_bow} and Figure~\ref{fig:exif_bow_approx}, the use of metadata reduces the execution time of K-Means
to the point where \emph{SIFT} become the bottleneck.  Improvements to the feature extraction step could further drive down the execution time of the pipeline.

%\textbf {Onboard GPU}
%GPUs could also provide significant improvement in performance and energy consumption of this pipeline. K-Means
%operations could be parallized across a GPU core since the most expensive step is highly parallelizable.  However, there may be implementation challenges
%and bottlenecks associated with copying data and objects from radio memory to CPU memory to GPU memory~\cite{gpunet},
%making the exact performance improvement unclear.  We leave further exploration of this optimization for future work.

%\input{pshift}
%\section{Contributions}
%In previous work, k-means is used to reduce sensor measurement error~\cite{uav_kmeans}.  
%While others use mathematical techniques to reduce the runtime at the cost of accuracy through approximation
%techniques~\cite{msr_approx_kmeans}.
%
%The contribution we make in this work is to use the onboard sensors on the mobile phone to speed up
%clustering computations.  We use contextual data to inform the clustering process.

\section{Discussion}
Today's edge computing frameworks, from IoT sensor devices to mobile phones are heavily reliant on
the cloud.  Most application architectures use the cloud for either offloading their computation or 
acting as a mediary between devices, altogether.  For high-density co-location scenarios, such as a 
stadium event or protest, the connection to the cloud through the cellular network is unstable or often
unavailable.  More generally, the growth in the number of cloud-reliant devices through the spread of IoT
and the explosive use of smart phones, strongly suggests that cloud-reliant architectures 
will not always be feasible, especially 
for applications where predictable quality of service and latency is important, cloud-based architectures
cannot guarantee either~\cite{cloud_not_enough}.  
%We assert that distributed computation at the edge is under-explored and believe that there is a tremendous 
%opportunity to enable a new class of applications that are less reliant on cloud infrastructure.
Our initial results suggest that computationally expensive pipelines can be executed at the edge, entirely,
and yield comparable performance through a combination of approximation techniques and contextual information.
Moving forward, we believe there should be further exploration on architectures and techniques that achieve
comparable performance at the edge to jobs executed in the cloud.

% Architectures for edge-based applications are heavily reliant on the cloud

% There are situations where that's not the best choice

% It's also not the best case moving forward more generally -- trends, etc.

% New approaches that can run at the edge must be explored and more research must be done about their feasibility

% Mini model and framework proposal.
%\subsection{Machine Learning Framework For Edge Devices} 
\label{sec:pshift}
This does not however imply that the cloud should be entirely ignored. Instead, we believe that the cloud should be used more judiciously. Below, 
we examine the notion of dynamically shifting components of the pipeline between the cloud and
the mobile network.  We model our four-stage pipeline as a linear combination of the processing times on the cloud and in the mobile network and data transmission time. We further used representative workloads to estimate their average processing times on the cloud and the mobile network.
Equation~\ref{eq:tot_cost} shows the sum of the aggregate components that our model considers.
\vspace{-0.5 mm}
\begin{equation}
\label{eq:tot_cost}
cost = t^{cloud}_{compute} + t^{mobile}_{compute} + t_{tx}
\end{equation}
\vspace{-0.3 mm}
 We calculated the cost across several pipeline configurations, shifting stages back and forth between the cloud and the mobile network as the available bandwidth changes. 
%Table~\ref{tab:pshift_params} shows the set of parameters on their values in our simulation.  
%\emph{Due to space constraints, we omit the details of the model.}

\begin{figure}[h!]
                \centering
        \includegraphics[width=0.45\textwidth]{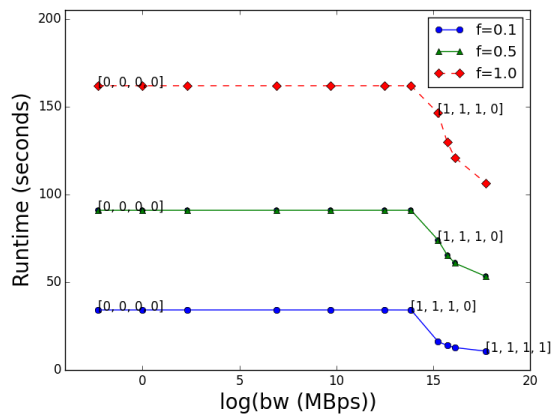}
                \caption{Model of pipeline shifting as a function of the available bandwidth.  Note that as more bandwidth becomes available, shifting
                    pipeline stages to the cloud yields better performance.}
                \label{fig:shift}
\end{figure}

A value of 0 corresponds to placement of that stage in the mobile network
and a 1 corresponds to placement in the cloud.   For example, $[ 0,0,0,0 ]$ means the entire pipeline runs in the mobile network while 
$[ 1,1,1,1 ]$ means the entire pipeline runs in the cloud.  Note, we only consider cases when the complete stage of a pipeline is shifted, i.e., 
pipeline shifting corresponds to a windowed placement shift for stages between the cloud and the mobile.  Therefore we consider only five 
configurations ranging from entirely 
in the mobile network to entirely in the cloud.  We also consider a different fraction of the data to send back by varying the fraction, $f$, of 
photos to transmit.
Figure~\ref{fig:shift} shows the results of our simulation.
We observe that shifting components of the pipeline could result in better performance.  For future work we look to explore
dynamic pipeline shifting in the context of an execution framework.

\subsection{GPUs, Mobility, and Failure}
There are several challenges that need to be addressed before we are able to fully realize the distributed computing potential of the high 
density of mobile phones. It is important to study the effects of mobility and intermittent connectivity between these phones which will in turn 
effect the distribution of the tasks. We need to design robust protocols that will be able to handle these lower-level failures that are 
characteristics of the mobile networks but still being able to support higher-level synchronous tasks. 
%Finally, can we also approximate answers from unresponsive phones?  
%Moreoever, as it becomes easier
%to use onboard GPUs on phones, how does that affect distributed machine learning jobs?  Can we make use of GPUs effectively in this environment?
In addition, there is also a constant increase in the computing power of these phones. GPUs could provide significant improvement in 
performance and energy consumption. However, there are implementation challenges
and bottlenecks associated with copying data and objects from radio memory to CPU memory to GPU memory%~\cite{gpunet},   
making the exact performance improvement unclear.  We leave further exploration of this optimization for future work.

% \balance
\bibliographystyle{abbrv}
\footnotesize
\bibliography{sigproc}  % sigproc.bib is the name of the Bibliography in this case

\begin{thebibliography}{10}

\bibitem{android_NDK}
Android ndk.
\newblock \url{https://developer.android.com/ndk/index.html}.

\bibitem{google_maps}
The bright side of sitting in traffic: Crowdsourcing road congestion data.
\newblock
  \url{https://googleblog.blogspot.com/2009/08/bright-side-of-sitting-in-traffic.html}.

\bibitem{exif}
Exchangeable image file format for digital still cameras: Exif version 2.2.
\newblock \url{http://www.exif.org/Exif2-2.PDF}.

\bibitem{pocket_cluster}
A cluster in your pocket:cell phone processors are getting more powerful. is a
  cell phone cluster possible?
\newblock {\em Linux J.}, 2010(196), 2010.

\bibitem{cisco_projection}
{Cisco Visual Networking Index: Global Mobile Data Traffic Forecast Update
  2014-2019}.
\newblock Technical report, Cisco, 01 2015.

\bibitem{surf}
H.~Bay, A.~Ess, T.~Tuytelaars, and L.~Van~Gool.
\newblock Speeded-up robust features (surf).
\newblock {\em Comput. Vis. Image Underst.}, 110(3):346--359, June 2008.

\bibitem{common_pipeline}
G.~Csurka, C.~R. Dance, L.~Fan, J.~Willamowski, and C.~Bray.
\newblock Visual categorization with bags of keypoints.
\newblock In {\em In Workshop on Statistical Learning in Computer Vision,
  ECCV}, pages 1--22, 2004.

\bibitem{codeoffload_longlast}
E.~Cuervo, A.~Balasubramanian, D.~ki~Cho, A.~Wolman, S.~Saroiu, R.~Ch, and
  P.~Bahl.
\newblock Maui: Making smartphones last longer with code offload.
\newblock In {\em In In Proceedings of ACM MobiSys}, 2010.

\bibitem{c1}
S.~Deng, L.~Huang, J.~Taheri, and A.~Zomaya.
\newblock Computation offloading for service workflow in mobile cloud
  computing.
\newblock {\em Parallel and Distributed Systems, IEEE Transactions on},
  PP(99):1--1, 2014.

\bibitem{fei2005bayesian}
L.~Fei-Fei and P.~Perona.
\newblock A bayesian hierarchical model for learning natural scene categories.
\newblock In {\em Computer Vision and Pattern Recognition, 2005. CVPR 2005.
  IEEE Computer Society Conference on}, volume~2, pages 524--531. IEEE, 2005.

\bibitem{activity_recog}
N.~D. Lane, Y.~Xu, H.~Lu, S.~Hu, T.~Choudhury, A.~T. Campbell, and F.~Zhao.
\newblock Enabling large-scale human activity inference on smartphones using
  community similarity networks (csn).
\newblock In {\em Proceedings of the 13th International Conference on
  Ubiquitous Computing}, UbiComp '11, pages 355--364, New York, NY, USA, 2011.
  ACM.

\bibitem{distr_kmeans}
Y.~Liang, M.~florina Balcan, and A.~Kanchanapally.
\newblock Distributed pca and k-means clustering.

\bibitem{kmeans}
S.~Lloyd.
\newblock Least squares quantization in pcm.
\newblock {\em Information Theory, IEEE Transactions on}, 28(2):129--137, Mar
  1982.

\bibitem{sift}
D.~G. Lowe.
\newblock Distinctive image features from scale-invariant keypoints.
\newblock {\em Int. J. Comput. Vision}, 60(2):91--110, Nov. 2004.

\bibitem{speaker_recog}
H.~Lu, A.~J.~B. Brush, B.~Priyantha, A.~K. Karlson, and J.~Liu.
\newblock Speakersense: Energy efficient unobtrusive speaker identification on
  mobile phones.
\newblock In {\em Proceedings of the 9th International Conference on Pervasive
  Computing}, Pervasive'11, pages 188--205, Berlin, Heidelberg, 2011.
  Springer-Verlag.

\bibitem{kmeans_nphard}
M.~Mahajan, P.~Nimbhorkar, and K.~Varadarajan.
\newblock The planar k-means problem is np-hard.
\newblock In {\em Proceedings of the 3rd International Workshop on Algorithms
  and Computation}, WALCOM '09, pages 274--285, Berlin, Heidelberg, 2009.
  Springer-Verlag.

\bibitem{darwin_phones}
E.~Miluzzo, C.~T. Cornelius, A.~Ramaswamy, T.~Choudhury, Z.~Liu, and A.~T.
  Campbell.
\newblock Darwin phones: The evolution of sensing and inference on mobile
  phones.
\newblock In {\em Proceedings of the 8th International Conference on Mobile
  Systems, Applications, and Services}, MobiSys '10, pages 5--20, New York, NY,
  USA, 2010. ACM.

\bibitem{emotion_detect}
K.~K. Rachuri, M.~Musolesi, C.~Mascolo, P.~J. Rentfrow, C.~Longworth, and
  A.~Aucinas.
\newblock Emotionsense: A mobile phones based adaptive platform for
  experimental social psychology research.
\newblock In {\em Proceedings of the 12th ACM International Conference on
  Ubiquitous Computing}, UbiComp '10, pages 281--290, New York, NY, USA, 2010.
  ACM.

\bibitem{mobile_transportation}
S.~Reddy, M.~Mun, J.~Burke, D.~Estrin, M.~Hansen, and M.~Srivastava.
\newblock Using mobile phones to determine transportation modes.
\newblock {\em ACM Trans. Sen. Netw.}, 6(2):13:1--13:27, Mar. 2010.

\bibitem{orb}
E.~Rublee, V.~Rabaud, K.~Konolige, and G.~Bradski.
\newblock Orb: An efficient alternative to sift or surf.
\newblock In {\em Computer Vision (ICCV), 2011 IEEE International Conference
  on}, pages 2564--2571, Nov 2011.

\bibitem{mobile_advert}
J.~P. Rula, B.~Jun, and F.~Bustamante.
\newblock Mobile ad(d): Estimating mobile app session times for better ads.
\newblock In {\em Proceedings of the 16th International Workshop on Mobile
  Computing Systems and Applications}, HotMobile '15, pages 123--128, New York,
  NY, USA, 2015. ACM.

\bibitem{cell_perf}
R.~Schoenen, A.~Otyakmaz, and Z.~Xu.
\newblock Resource allocation and scheduling in fdd multihop cellular systems.
\newblock In {\em Communications Workshops, 2009. ICC Workshops 2009. IEEE
  International Conference on}, pages 1--6, June 2009.

\bibitem{mobile_human_queue}
Y.~Wang, J.~Yang, Y.~Chen, H.~Liu, M.~Gruteser, and R.~P. Martin.
\newblock Tracking human queues using single-point signal monitoring.
\newblock In {\em Proceedings of the 12th Annual International Conference on
  Mobile Systems, Applications, and Services}, MobiSys '14, pages 42--54, New
  York, NY, USA, 2014. ACM.

\bibitem{kmapprox}
G.~Zeng.
\newblock Fast approximate k-means via cluster closures.
\newblock In {\em Proceedings of the 2012 IEEE Conference on Computer Vision
  and Pattern Recognition (CVPR)}, CVPR '12, pages 3037--3044, Washington, DC,
  USA, 2012. IEEE Computer Society.

\bibitem{cloud_not_enough}
B.~Zhang, N.~Mor, J.~Kolb, D.~S. Chan, K.~Lutz, E.~Allman, J.~Wawrzynek,
  E.~Lee, and J.~Kubiatowicz.
\newblock The cloud is not enough: Saving iot from the cloud.
\newblock In {\em 7th USENIX Workshop on Hot Topics in Cloud Computing
  (HotCloud 15)}, Santa Clara, CA, July 2015. USENIX Association.

\end{thebibliography}
\end{document}